\documentclass[conference,10pt]{IEEEtran}
\IEEEoverridecommandlockouts

\usepackage{cite}
\usepackage{makecell}
\usepackage{float}
\usepackage{amsmath,amssymb,amsfonts}
\usepackage{algorithmic}
\usepackage{graphicx}
\usepackage{textcomp}
\usepackage{xcolor}
\usepackage{newtxtext}
\usepackage{newtxmath}
\usepackage{setspace}
\def\BibTeX{{\rm B\kern-.05em{\sc i\kern-.025em b}\kern-.08em
    T\kern-.1667em\lower.7ex\hbox{E}\kern-.125emX}}
\begin{document}

\title{Linguistic Indicators of Early Cognitive Decline in the DementiaBank Pitt Corpus: A Statistical and Machine Learning Study}

\author{\IEEEauthorblockN{Artsvik Avetisyan, Sachin Kumar}
\IEEEauthorblockA{Zaven P. and Sonia Akian College of Science \& Engineering, American University of Armenia, Yerevan, Armenia \\artsvik\_avetisyan@edu.aua.am, s.kumar@aua.am}}

\maketitle

\begin{abstract}
Background:
Subtle changes in spontaneous language production are among the earliest indicators of cognitive decline. Identifying linguistically interpretable markers of dementia can support transparent and clinically grounded screening approaches.

Methods:
This study analyzes spontaneous speech transcripts from the DementiaBank Pitt Corpus using three linguistic representations: raw cleaned text, a part-of-speech (POS) - enhanced representation combining lexical and grammatical information, and a POS-only syntactic representation. Logistic regression and random forest models were evaluated under two protocols: transcript-level train–test splits and subject-level five-fold cross-validation to prevent speaker overlap. Model interpretability was examined using global feature importance, and statistical validation was conducted using Mann–Whitney U tests with Cliff’s delta effect sizes.

Results:
Across representations, models achieved stable performance, with syntactic and grammatical features retaining strong discriminative power even in the absence of lexical content. Subject-level evaluation yielded more conservative but consistent results, particularly for POS-enhanced and POS-only representations. Statistical analysis revealed significant group differences in functional word usage, lexical diversity, sentence structure, and discourse coherence, aligning closely with machine learning feature importance findings.

Conclusion:
The results demonstrate that abstract linguistic features capture robust markers of early cognitive decline under clinically realistic evaluation. By combining interpretable machine learning with non-parametric statistical validation, this study supports the use of linguistically grounded features for transparent and reliable language-based cognitive screening.
\end{abstract}


\section{Introduction}

Early identification of cognitive decline is an important goal in clinical research and healthcare \cite{r1}. Neurodegenerative conditions such as Alzheimer's disease progress slowly, and the earliest signs often appear in spontaneous communication \cite{r2}. Subtle changes in spoken language, including reduced vocabulary, simplified sentence structure, increased hesitation, and difficulties maintaining coherent discourse, can emerge long before a formal diagnosis is made. Although these changes may be difficult for human listeners to detect reliably, they can be observed through systematic analysis of speech transcripts. This makes language a valuable source of information for understanding early cognitive changes and supports the growing interest in computational approaches to clinical linguistics \cite{r3}.

The DementiaBank Pitt Corpus provides a well-established foundation for studying these patterns \cite{r4}. It contains detailed CHAT-format transcripts of several diagnostic speech tasks performed by individuals with dementia and age-matched healthy controls. These tasks include picture description, story recall, verbal fluency, and sentence repetition. The transcripts preserve important linguistic details such as pauses, repetitions, fillers, and disfluencies, which are essential for examining how language behavior changes during cognitive decline. Because the corpus includes both spontaneous and structured speech, it allows researchers to investigate lexical, syntactic, and semantic aspects of language in a controlled and comprehensive manner \cite{r5}.

Previous work has shown that early dementia is often associated with shorter utterances, reduced syntactic complexity, limited lexical diversity, and more frequent fillers or repeated words \cite{r6,r7,r8}. While these findings provide useful insight, many earlier studies rely on a narrow selection of handcrafted features or single-task analyses. There remains limited understanding of how different levels of linguistic representation, such as raw text compared to part-of-speech patterns, influence the detection of cognitive impairment. In addition, there is a need to connect machine learning findings with statistical evidence to determine which linguistic features are most consistently associated with cognitive decline rather than being predictive only within a specific model \cite{r10}.

The present study addresses these gaps by analyzing linguistic features extracted from the Pitt Corpus in order to identify which characteristics of language are most strongly linked to early cognitive decline. We work with three forms of representation: raw cleaned text, text enriched with part-of-speech and content-word tags, and part-of-speech representations with content-word tags removed. Using these representations, we train logistic regression and random forest models to classify participants as either dementia or control. After training, we extract the most influential features from each model to support interpretability and to understand how different representations shape the classification process.

To strengthen the reliability of these findings, we complement the machine learning analysis with statistical tests and effect size calculations. This allows us to evaluate whether features highlighted by the models also show measurable differences between the dementia and control groups. By comparing statistical associations with model-based feature importance, we aim to identify linguistic features that consistently reflect early cognitive decline across both analytical perspectives.

This study offers a detailed and transparent examination of linguistic markers of cognitive decline using multiple feature representations and interpretable models. By combining machine learning with statistical validation, it provides a clearer understanding of how lexical, syntactic, and part-of-speech patterns may serve as early indicators of cognitive impairment.

\section{Related Work}

Research on language and cognitive decline has a long history in both clinical linguistics and cognitive science. Early studies showed that people in the initial stages of Alzheimer's disease often produce shorter and less complex sentences, rely more on high-frequency and generic words, and show difficulties in retrieving specific terms \cite{r10,r11}. These linguistic changes appear even when other cognitive abilities are still relatively preserved, which makes language an important window into early decline \cite{r12,r13}. Several clinical studies have also noted increased hesitation, repetitions, and reduced coherence in narrative speech, suggesting that both lexical and discourse-level features are affected \cite{r14}.
Reviews of spontaneous speech in cognitive decline consistently identify changes in functional word usage, lexical diversity, and discourse structure as early and robust indicators, even in preclinical or subjective stages \cite{r14.1}.

With the development of larger clinical speech datasets, such as the DementiaBank Pitt Corpus, computational methods have become common in this area. Much of the recent work focuses on extracting linguistic features from the Cookie Theft picture description task, where participants describe a fixed image \cite{r15}. This task has been widely analysed because it provides spontaneous yet comparable speech samples across participants. Studies using this task have reported reductions in lexical richness, fewer content words, simpler syntactic structures, and more frequent disfluencies in participants with dementia. Other work has explored narrative tasks and story recall, showing that individuals with cognitive impairment often omit key elements of a story or produce less organised narratives.

Machine learning approaches have also been applied to dementia detection from language \cite{r16}. Early work used lexical counts, syntactic statistics, and readability scores as features for traditional classifiers such as logistic regression and support vector machines \cite{r17}. More recent studies introduced part-of-speech features, dependency parses, and embeddings to capture more subtle linguistic patterns \cite{r17}. Although deep learning models such as RNNs and transformer-based architectures have achieved strong performance, many of these models offer limited interpretability, which is a challenge in clinical settings where explanations are important.

A number of studies have also explored feature importance and model explainability \cite{r18,r19}. Some work has used regression coefficients or decision tree weights to highlight influential features, while others have applied methods such as SHAP or LIME to interpret model predictions \cite{r20}. These attempts show that interpretable machine learning can help clinicians understand which aspects of language behavior differ most clearly between healthy and impaired speech. However, many studies do not combine model explainability with direct statistical testing, which can provide an independent check on whether a feature is meaningfully associated with cognitive decline.

There is also growing interest in comparing different representations of language. Some researchers work directly with raw words, while others focus on syntactic or part-of-speech patterns to isolate grammatical behavior from vocabulary effects \cite{r20}. Abstracting away from specific words can reveal deeper structural changes, but results vary across datasets and tasks \cite{r21}. More work is needed to understand how different levels of linguistic representation contribute to the detection of early cognitive decline.

Research indicates clear linguistic differences between healthy and cognitively impaired individuals, but the field still lacks studies that integrate multiple feature representations, interpretable models, and statistical validation \cite{r22}. This work builds on prior studies by examining raw text, POS-enhanced text, and POS-only representations on the same dataset, while combining machine learning feature importance with statistical association tests to provide a more comprehensive view of the linguistic features associated with early cognitive decline.

\section{Dataset Description}

This study uses the DementiaBank Pitt Corpus, one of the most widely used resources for analyzing language in Alzheimer’s disease and related neurocognitive conditions. The corpus contains transcripts of spoken language collected from individuals diagnosed with probable Alzheimer’s disease and from cognitively healthy older adults who serve as control participants. All transcripts follow the CHAT transcription format, which preserves important details such as fillers, pauses, repetitions, and other speech behaviors that are often relevant for studying cognitive decline. Table~\ref{tab:tab1} provides an overview of the Pitt Corpus used in this work.

\begin{table}[h] 
    \centering 
    \caption{DementiaBank Pitt Corpus} 
    \label{tab:tab1}
    \begin{tabular}{lc} \hline 
    Characteristic & Value \\ 
    \hline 
    Dataset & DementiaBank Pitt Corpus \\ 
    Task & Cookie Theft description \\ 
    Language & English \\ 
    Total transcripts & 500 \\ 
    Control transcripts & 243 \\ 
    Dementia transcripts & 257 \\ 
    \hline 
    \end{tabular} 
\end{table}

The dataset is organised into two main diagnostic groups: \textit{control} and \textit{dementia}. Each group contains transcripts from four different speech tasks, including Cookie Theft picture description, story recall, verbal fluency, and sentence repetition. These tasks capture different aspects of language production. The Cookie Theft task elicits spontaneous narrative speech, the story recall task assesses memory for structured content, the verbal fluency task probes lexical retrieval under time constraints, and the sentence repetition task evaluates syntactic processing and working memory. Together, these tasks allow for a broad examination of linguistic behavior across both spontaneous and structured speaking contexts.

All transcripts are stored as \texttt{.cha} files and include speaker labels as well as multiple annotation tiers. For the purposes of this study, only utterances produced by the participant were analysed. Interviewer speech was removed, and transcription markers unrelated to linguistic content were cleaned as needed. Pauses, fillers, and repetitions were retained when relevant, as these phenomena may reflect linguistic and cognitive processes affected by dementia.

The participants in the dementia group were diagnosed with probable Alzheimer’s disease, while the control group consisted of older adults without diagnosed cognitive impairment. The corpus includes multiple recordings from some participants collected across different sessions. Because this longitudinal structure can influence evaluation outcomes, we adopt two complementary experimental protocols in this study. First, we report transcript-level baseline experiments in which individual transcripts are treated as separate samples, consistent with common practice in prior work. Second, to obtain a more conservative and clinically realistic estimate of generalization, we conduct subject-level evaluation using grouped cross-validation, ensuring that all transcripts from a given participant are assigned to the same fold and that no individual appears in both training and testing data.

The number of available transcripts varies slightly across tasks, as not all participants completed every task. Nevertheless, both diagnostic groups contain a sufficient number of samples to support reliable statistical analysis and machine learning experiments. The availability of parallel tasks for dementia and control participants, combined with subject-level evaluation, makes the Pitt Corpus suitable for examining lexical, syntactic, and part-of-speech-based features while accounting for inter-speaker variability.

\section{Methodology}

\subsection{Feature Sets}

To examine how different linguistic representations influence the identification of early cognitive decline, three feature sets were constructed from the transcript data. Each representation captures a distinct level of linguistic information, allowing us to compare the contribution of lexical content, syntactic structure, and their combination.

The first feature set is based on cleaned textual data. After preprocessing, which included removing interviewer utterances and non-linguistic transcription markers, participant speech was represented using text-derived linguistic features computed from tokenized transcripts. This representation preserves lexical choice and reflects how much discriminative information can be obtained from word usage patterns alone, without explicit modeling of grammatical structure.

The second feature set combines lexical and syntactic information by augmenting tokens with part-of-speech labels. In this representation, content words such as nouns, verbs, adjectives, and adverbs were retained in their original lexical form, while function words were represented using their grammatical categories. This design allows the model to jointly capture semantic content and syntactic organization, while reducing sensitivity to surface-level variation in high-frequency function words.

The third feature set focuses exclusively on syntactic structure. In this case, lexical content was removed and tokens were represented solely by their part-of-speech tags. Content-word tags were excluded so that the representation reflects grammatical patterns rather than semantic information. This abstraction allows us to investigate whether structural properties of language, independent of lexical meaning, provide sufficient signals to distinguish between dementia and control speech.

\subsection{Machine Learning Models}

Two supervised machine learning models were used in this study: logistic regression and random forest. These models were selected because they are well-established, computationally efficient, and offer a high degree of interpretability, which is particularly important in clinical language analysis and biomedical applications.

Logistic regression was used as a linear baseline model to estimate the probability that a transcript belongs to the dementia group given a vector of linguistic features. The conditional probability is defined as
\[
P(y = 1 \mid x) = \frac{1}{1 + e^{-(\beta_0 + \beta^T x)}},
\]
where $x$ represents the extracted feature vector, $\beta$ denotes the learned coefficients, and $\beta_0$ is the bias term. To improve robustness and reduce overfitting, an L2 regularization term was applied, and class imbalance was addressed using class-weighted loss. All input features were standardized prior to training. The learned coefficients provide direct insight into the direction and relative importance of individual linguistic features, making logistic regression well suited for interpretability-focused analysis.

The random forest was employed as a non-linear ensemble model to capture more complex relationships among linguistic features. The model consists of an ensemble of decision trees trained on bootstrap samples of the data, with random feature subsampling at each split. Class balancing was applied to account for unequal group sizes. The random forest complements logistic regression by modeling non-linear interactions between linguistic features that may not be adequately captured by linear decision boundaries, while still allowing global feature importance analysis.

Hyperparameters for both models were explored within standard ranges reported in prior work using training folds only, after which the final configurations were fixed across all reported evaluations to avoid fold-specific optimization. This design choice avoids overly aggressive tuning and reduces the risk of overfitting on a moderate-sized clinical dataset, while still allowing the models to capture meaningful linguistic patterns.

\subsection{Feature Importance Extraction}

To support interpretability, feature importance was extracted from both classification models using global, model-specific criteria. Our analysis focuses on identifying linguistic features that consistently contribute to model decisions, rather than providing local or instance-level explanations.

For logistic regression, feature importance was quantified using the absolute magnitude of the learned coefficients. Features with larger absolute coefficients exert a stronger influence on the model’s decision boundary, with the sign of the coefficient indicating the direction of association with the dementia or control class. For each feature representation, the top twenty features were selected based on their absolute coefficient values. When subject-level cross-validation was used, coefficients were aggregated across folds to assess the stability of feature importance rankings.

For the random forest model, feature importance was computed based on the mean decrease in impurity across all trees in the ensemble. This metric reflects how frequently and effectively a feature contributes to reducing classification uncertainty during tree construction. The top twenty features with the highest importance scores were extracted for further analysis, and importance values were averaged across cross-validation folds when applicable.

Comparing feature rankings across logistic regression and random forest models allows us to examine whether certain linguistic features consistently emerge as informative under both linear and non-linear modeling assumptions. While this approach provides a global view of feature relevance, it does not capture local explanation fidelity or user-centered interpretability. Accordingly, the reported feature importance results should be interpreted as descriptive indicators of model behavior rather than clinically validated explanations.

\subsection{Statistical Association Analysis}

In addition to machine learning analysis, non-parametric statistical tests were conducted to independently examine the association between linguistic features and cognitive decline. For each feature, distributions were compared between the dementia and control groups to assess whether systematic group-level differences were present.

Because many linguistic features exhibited non-normal distributions, group comparisons were performed using the Mann–Whitney U test, which does not assume normality and is well suited for heterogeneous clinical data. This test evaluates whether values from one group tend to be systematically higher or lower than values from the other group.

To quantify the magnitude and direction of group differences, Cliff’s delta was used as a non-parametric effect size measure. Given two samples $A = \{a_1, \dots, a_m\}$ and $B = \{b_1, \dots, b_n\}$, Cliff’s delta is defined as
\[
\delta = \frac{\sum_{i=1}^{m} \sum_{j=1}^{n} \operatorname{sgn}(a_i - b_j)}{m \times n},
\]
where $\operatorname{sgn}(\cdot)$ denotes the sign function. The resulting value ranges from $-1$ to $1$, with positive values indicating that feature values tend to be higher in the control group and negative values indicating higher values in the dementia group. Unlike parametric effect sizes, Cliff’s delta does not rely on distributional assumptions and provides a robust measure of effect strength for skewed or heavy-tailed data.

When multiple features were tested simultaneously, p-values were adjusted using the Benjamini–Hochberg procedure to control the false discovery rate. This statistical analysis complements the machine learning results by identifying linguistic features that exhibit statistically meaningful group differences independent of any specific classifier. Together, these analyses help distinguish features that are consistently associated with cognitive decline from those that are predictive only within a particular modeling framework.

\section{Results}
The results are reported under two experimental protocols. First, transcript-level train–test splits are presented as baseline results to facilitate comparison with prior studies that treat individual transcripts as independent samples. Second, subject-level cross-validation results are reported to ensure that no participant appears in both training and testing sets, providing a more conservative and clinically realistic evaluation.

\subsection{Baseline Transcript-Level Results}

\subsubsection{Classification Performance on Raw Cleaned Text}

The first set of experiments was conducted using raw cleaned textual features extracted from the Pitt Corpus transcripts. These features included measures of lexical richness, syntactic structure, and semantic coherence derived directly from participant speech. The dataset was divided into training and test sets using an 80/20 stratified split at the transcript level to preserve the class distribution between dementia and control groups.

Table~\ref{tab:tab2} reports the classification performance of logistic regression and random forest models on the held-out test set. Logistic regression achieved an overall accuracy of 0.68 and demonstrated relatively balanced performance across the two classes. For the dementia group, the model achieved a precision of 0.75, a recall of 0.66, and an F1-score of 0.70. Performance on the control group was slightly lower, with an F1-score of 0.67. The macro-averaged F1-score of 0.68 indicates stable performance across both classes.

The random forest model achieved an overall accuracy of 0.66. It showed higher recall for the dementia class at 0.74, but lower recall for the control class, resulting in a control F1-score of 0.59. The dementia class F1-score was 0.71. The macro-averaged F1-score for the random forest model was 0.65. Although the overall accuracy was slightly lower than that of logistic regression, the random forest model demonstrated stronger sensitivity to dementia cases.

\begin{table}[h]
\centering
\caption{Classification performance on raw cleaned text features}
\label{tab:tab2}
\begin{tabular}{lcccc}
\hline
Model & Accuracy & Precision & Recall & F1-score \\
\hline
Logistic Regression & 0.68 & 0.69 & 0.69 & 0.68 \\
Random Forest & 0.66 & 0.66 & 0.66 & 0.65 \\
\hline
\end{tabular}
\end{table}

\subsubsection{Feature Importance Analysis}

\begin{figure}[h]
\centering
\includegraphics[width=0.8\linewidth]{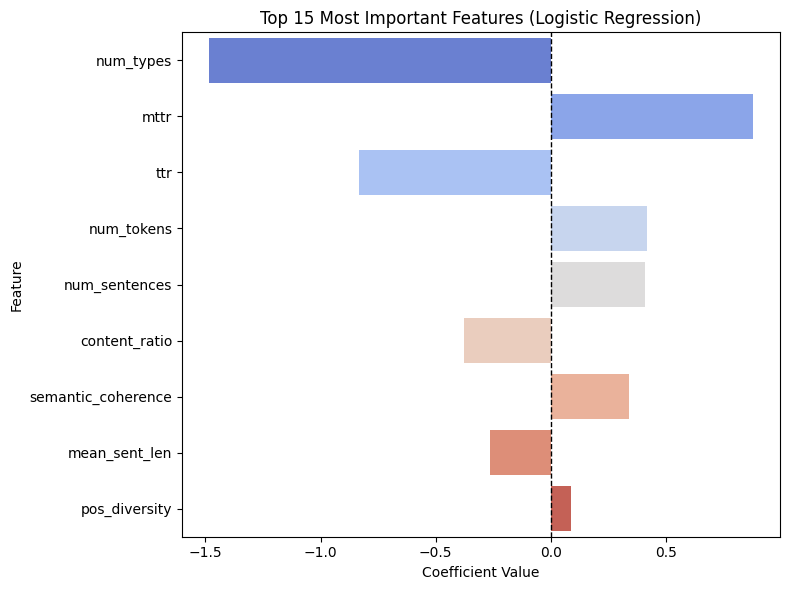}
\caption{Top logistic regression features for the raw cleaned text representation}
\label{fig:fig1}
\end{figure}

To examine which linguistic features contributed most strongly to classification decisions, feature importance was extracted from both machine learning models. For logistic regression, importance was determined using the magnitude and direction of the learned coefficients after feature scaling. Figure~\ref{fig:fig1} shows the most influential features identified by the logistic regression model trained on raw cleaned text.

The logistic regression model highlighted lexical diversity measures as particularly important. Features such as the number of unique word types, type-token ratio, and moving-average type-token ratio showed strong associations with classification outcomes. Structural measures including the number of tokens, number of sentences, and mean sentence length were also influential. In addition, discourse-level features such as semantic coherence and the ratio of content words contributed to the model’s decisions. The direction of the coefficients indicates that higher lexical diversity, longer sentences, and greater semantic coherence were more strongly associated with control transcripts, while reduced diversity and simpler, less coherent speech were associated with dementia transcripts.

\begin{figure}[h]
\centering
\includegraphics[width=0.8\linewidth]{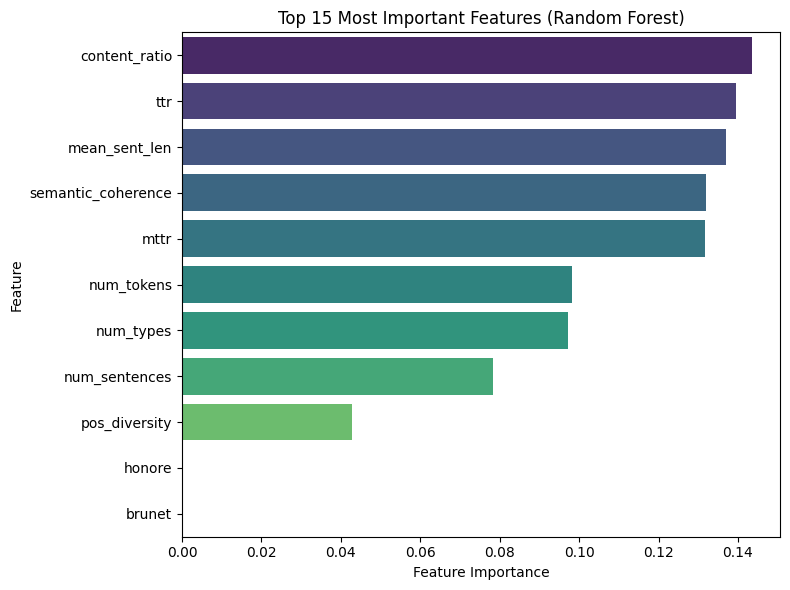}
\caption{Top random forest features for the raw cleaned text representation}
\label{fig:fig2}
\end{figure}

Feature importance analysis for the random forest model is shown in Figure~\ref{fig:fig2}. The most important feature identified by the random forest model was the content word ratio, followed by lexical diversity measures such as type-token ratio and moving-average type-token ratio. Measures of sentence structure, including mean sentence length and number of sentences, also ranked highly. Semantic coherence appeared among the top predictors, indicating that discourse-level organization plays a role in distinguishing between the two groups. Speech production measures, such as the number of tokens and word types, also contributed to the model’s performance.

Although the exact ranking of features differed between logistic regression and random forest, there was substantial overlap in the types of features identified as important. Both models consistently emphasized lexical diversity, sentence-level structure, and semantic coherence. This convergence across linear and non-linear models suggests that these features capture stable linguistic differences between dementia and control speech rather than model-specific artifacts.

\subsubsection{Classification Performance with POS-Enhanced Representation}

The second set of experiments was conducted using a POS-enhanced representation in which raw text features were combined with part-of-speech information and content-word tags. In this representation, content words were retained in their lexical form, while grammatical and functional information was captured through POS tags. This approach was designed to preserve semantic content while making syntactic patterns more explicit.

Using the same 80/20 stratified train-test split as in the raw-text experiments, logistic regression achieved an accuracy of 0.75 on the held-out test set. Performance was balanced across both classes, with an F1-score of 0.74 for the control group and 0.75 for the dementia group. The macro-averaged F1-score was 0.75, indicating improved and more stable performance compared to the raw-text representation. Precision for the dementia class increased to 0.83, while recall remained at 0.69, suggesting better discrimination of dementia-related linguistic patterns (Table~\ref{tab:tab3}).

The random forest model trained on the POS-enhanced representation achieved an accuracy of 0.72. The model produced an F1-score of 0.75 for the dementia class and 0.68 for the control class, with a macro-averaged F1-score of 0.72. Compared to the raw-text experiments, both models showed improved performance, indicating that the inclusion of syntactic information positively contributes to classification accuracy.

\begin{table}[h]
\centering
\caption{Classification performance using POS-enhanced features}
\label{tab:tab3}
\begin{tabular}{lcccc}
\hline
Model & Accuracy & Precision & Recall & F1-score \\
\hline
Logistic Regression & 0.75 & 0.76 & 0.75 & 0.75 \\
Random Forest & 0.72 & 0.72 & 0.72 & 0.72 \\
\hline
\end{tabular}
\end{table}

\subsubsection{Feature Importance Analysis for POS-Enhanced Representation}

\begin{figure}[h]
\centering
\includegraphics[width=0.8\linewidth]{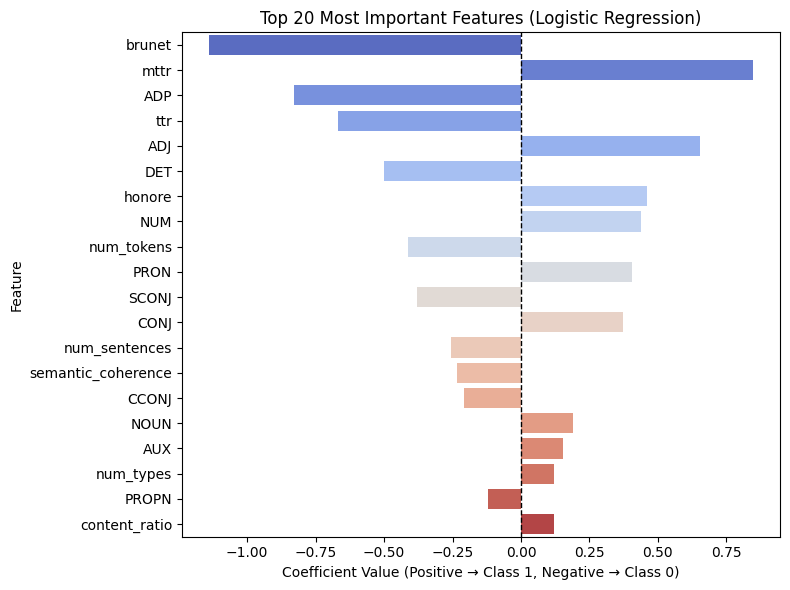}
\caption{Top logistic regression features for the POS-enhanced representation}
\label{fig:fig3}
\end{figure}

Feature importance analysis was performed to examine how lexical and syntactic features contributed to classification decisions in the POS-enhanced setting. For logistic regression, importance was determined based on the magnitude and direction of the learned coefficients. Figure~\ref{fig:fig3} shows the most influential features identified by the logistic regression model.

The logistic regression model highlighted a combination of lexical richness measures and part-of-speech features. Measures such as moving-average type-token ratio and type-token ratio remained among the strongest predictors, confirming the importance of lexical diversity. Several POS-based features, including adjectives, determiners, pronouns, conjunctions, and auxiliary verbs, also ranked highly. This indicates that grammatical usage patterns, in addition to vocabulary richness, play an important role in distinguishing between dementia and control speech. Semantic coherence and sentence-level measures such as the number of sentences continued to contribute, although with lower relative importance.

\begin{figure}[h]
\centering
\includegraphics[width=0.8\linewidth]{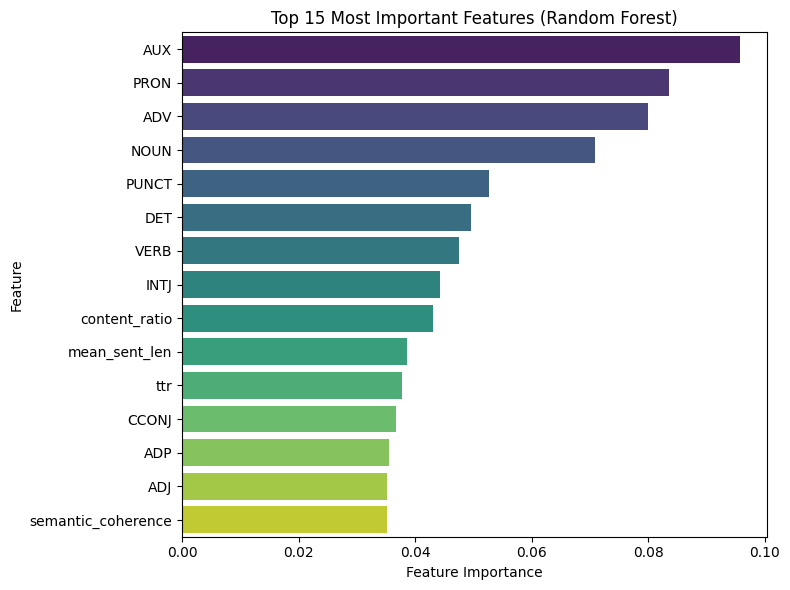}
\caption{Top random forest features for the POS-enhanced representation}
\label{fig:fig4}
\end{figure}

The random forest feature importance results for the POS-enhanced representation are shown in Figure~\ref{fig:fig4}. In contrast to logistic regression, the random forest model placed greater emphasis on POS-based features. Auxiliary verbs, pronouns, adverbs, nouns, and punctuation emerged as the most influential predictors. Content word ratio and mean sentence length also appeared among the top features, suggesting that both grammatical structure and speech production characteristics contribute to classification performance. Lexical diversity measures such as type-token ratio remained relevant but were ranked lower than several POS features.

Despite differences in ranking between the two models, both consistently identified a combination of lexical diversity, grammatical structure, and discourse-level features as important. Compared to the raw-text experiments, the POS-enhanced representation resulted in a clearer separation between dementia and control transcripts, particularly by emphasizing syntactic and functional word usage patterns. This suggests that incorporating part-of-speech information helps reveal linguistic characteristics associated with early cognitive decline that may not be fully captured by lexical content alone.

\subsubsection{Classification Performance with POS-Only Representation}

The final set of experiments was conducted using the POS-only representation, in which lexical content was removed and transcripts were represented exclusively through part-of-speech tags and structural linguistic features. This representation was designed to evaluate whether syntactic and grammatical patterns alone provide sufficient information to distinguish between dementia and control speech.

Using the same 80/20 stratified train-test split as in previous experiments, logistic regression achieved an improved accuracy of 0.75 on the held-out test set. The model demonstrated strong recall for the control group at 0.82, while recall for the dementia group was 0.69. Precision was higher for the dementia class at 0.83 compared to 0.68 for the control class. As a result, the F1-scores were 0.74 for the control group and 0.75 for the dementia group, yielding a macro-averaged F1-score of 0.75. These results indicate stable and balanced classification performance despite the absence of lexical information.

The random forest model achieved a similar accuracy of 0.75. Recall for the dementia class increased to 0.79, while recall for the control class was 0.69. Precision values were 0.77 for the dementia group and 0.72 for the control group, resulting in F1-scores of 0.78 and 0.71, respectively. The macro-averaged F1-score for the random forest model was 0.74. Compared to logistic regression, the random forest model showed slightly higher sensitivity to dementia-related patterns, although overall performance remained comparable.

Table~\ref{tab:tab4} reports the classification performance of both models using the POS-only representation. Performance remained comparable to that achieved with the POS-enhanced features and exceeded that of the raw-text representation, indicating that grammatical structure and part-of-speech distributions alone capture a substantial portion of the linguistic signal associated with early cognitive decline.

\begin{table}[h]
\centering
\caption{Classification performance using POS-only features}
\label{tab:tab4}
\begin{tabular}{lcccc}
\hline
Model & Accuracy & Precision & Recall & F1-score \\
\hline
Logistic Regression & 0.75 & 0.75 & 0.75 & 0.75 \\
Random Forest & 0.75 & 0.74 & 0.74 & 0.74 \\
\hline
\end{tabular}
\end{table}

\subsubsection{Feature Importance Analysis for POS-Only Representation}

\begin{figure}[h]
\centering
\includegraphics[width=0.9\linewidth]{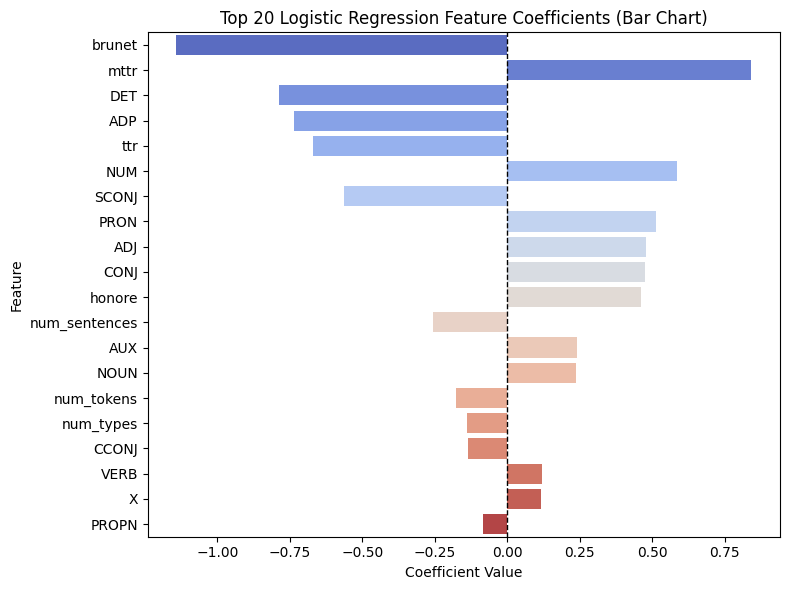}
\caption{Top logistic regression feature coefficients for the POS-only representation. Positive coefficients indicate features associated with dementia transcripts, while negative coefficients are associated with control transcripts.}
\label{fig:fig5}
\end{figure}

To examine which syntactic and grammatical features contributed most strongly to classification decisions, feature importance was extracted from both models trained on the POS-only representation. For logistic regression, importance was assessed using the magnitude and direction of the learned coefficients. The most influential features identified by this model are shown in Figure~\ref{fig:fig5}.

The logistic regression model highlighted several part-of-speech categories as strong predictors. Determiners, adpositions, subordinating conjunctions, pronouns, and numerals emerged among the most influential features, indicating that functional word usage and grammatical relationships play a central role in distinguishing between the two groups. Measures related to lexical diversity, including moving-average type-token ratio and type-token ratio, also retained influence despite the removal of lexical identity. Structural measures such as the number of sentences and total number of tokens further contributed to classification, reflecting differences in speech organization and production.

\begin{figure}[h]
\centering
\includegraphics[width=0.9\linewidth]{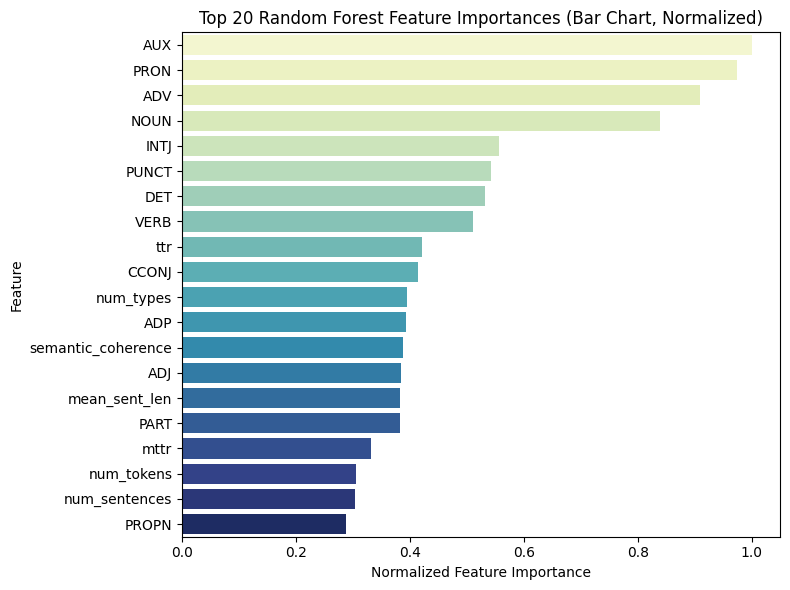}
\caption{Top random forest feature importances for the POS-only representation. Importance scores are normalized and reflect the contribution of each feature to reducing classification impurity across the ensemble.}
\label{fig:fig6}
\end{figure}

Feature importance results for the random forest model are shown in Figure \ref{fig:fig6}. In contrast to logistic regression, the random forest model placed greater emphasis on functional and grammatical categories. Auxiliary verbs and pronouns emerged as the most influential features, followed by adverbs and nouns. Disfluency-related markers, including interjections and punctuation, also ranked highly, highlighting their role in capturing irregularities in speech flow. Structural features such as sentence length and token counts contributed moderately, while semantic coherence appeared with lower relative importance compared to earlier representations.

Across both models, there was substantial agreement in the types of features identified as important. Pronouns, auxiliary verbs, conjunctions, and determiners consistently emerged as key indicators of group differences. These findings align with clinical observations that individuals with cognitive decline often exhibit difficulties with reference, tense marking, and sentence planning. The convergence of results across linear and non-linear models supports the robustness of syntactic and grammatical features as indicators of early cognitive decline.

The POS-only analysis shows that abstract grammatical patterns capture meaningful linguistic differences between dementia and control speech. While lexical information adds further interpretability, the findings indicate that degradation in syntactic structure and functional word usage is a core linguistic feature of early cognitive impairment, detectable even in highly abstracted representations.

\subsubsection{Statistical Association Analysis of Linguistic Features}
\begin{table*}[h]
\centering
\caption{Top linguistic features associated with cognitive decline based on statistical tests and effect sizes}
\label{tab:tab5}
\begin{tabular}{lccccc}
\hline
Feature & Mean Control & Mean Dementia & Cliff’s $\delta$ & $p$-value & $p_{adj}$ \\
\hline
ADV & 0.0248 & 0.0412 & -0.41 & $1.0\times10^{-16}$ & $2.9\times10^{-15}$ \\
PRON & 0.1139 & 0.1395 & -0.39 & $2.9\times10^{-15}$ & $4.1\times10^{-14}$ \\
NOUN & 0.1811 & 0.1558 & 0.37 & $4.2\times10^{-14}$ & $3.9\times10^{-13}$ \\
AUX & 0.0700 & 0.0559 & 0.35 & $8.4\times10^{-13}$ & $5.9\times10^{-12}$ \\
DET & 0.1201 & 0.1054 & 0.30 & $7.6\times10^{-10}$ & $4.3\times10^{-9}$ \\
INTJ & 0.0239 & 0.0327 & -0.30 & $9.0\times10^{-10}$ & $4.2\times10^{-9}$ \\
PUNCT & 0.1218 & 0.1329 & -0.24 & $9.4\times10^{-7}$ & $3.8\times10^{-6}$ \\
Mean sentence length & 7.83 & 7.25 & 0.18 & $2.2\times10^{-4}$ & $7.8\times10^{-4}$ \\
TTR & 0.5797 & 0.5510 & 0.18 & $4.0\times10^{-4}$ & $1.2\times10^{-3}$ \\
ADJ & 0.0303 & 0.0261 & 0.16 & $1.3\times10^{-3}$ & $3.8\times10^{-3}$ \\
ADP & 0.0869 & 0.0817 & 0.14 & $5.9\times10^{-3}$ & $1.5\times10^{-2}$ \\
Semantic coherence & 0.235 & 0.250 & -0.14 & $6.1\times10^{-3}$ & $1.4\times10^{-2}$ \\
Number of sentences & 16.59 & 18.82 & -0.12 & $1.4\times10^{-2}$ & $3.1\times10^{-2}$ \\
PROPN & 0.0134 & 0.0155 & -0.10 & $4.7\times10^{-2}$ & $9.4\times10^{-2}$ \\
\hline
\end{tabular}
\end{table*}

In addition to machine learning analysis, non-parametric statistical tests were conducted to identify linguistic features that were independently associated with early cognitive decline. For each numeric feature, distributions were compared between dementia and control transcripts using the Mann–Whitney U test, as feature values did not consistently satisfy normality assumptions. Effect sizes were quantified using Cliff’s delta, a non-parametric measure that captures the probability that a randomly selected observation from one group exceeds a randomly selected observation from the other. To account for multiple comparisons, p-values were adjusted using the Benjamini–Hochberg procedure.

Table~\ref{tab:tab5} presents the top linguistic features ranked by the absolute magnitude of Cliff’s delta. Several part-of-speech categories exhibited moderate to strong and statistically significant group differences. Adverbs and pronouns showed the largest negative Cliff’s delta values, indicating higher relative usage in dementia transcripts compared to control transcripts. In contrast, nouns, auxiliary verbs, and determiners displayed positive Cliff’s delta values, reflecting higher relative proportions in control speech. These patterns suggest that dementia-related language is characterized by increased reliance on certain functional and modifier categories alongside reduced use of content-bearing grammatical constructions.

Disfluency-related and interactional markers also demonstrated meaningful group differences. Interjections and punctuation occurred more frequently in dementia transcripts, consistent with increased hesitation, disrupted speech flow, and irregular turn construction. Lexical and structural measures further differentiated the groups. Type-token ratio and mean sentence length were higher in control transcripts, whereas the number of sentences was greater in dementia transcripts, reflecting shorter, more fragmented utterances. Semantic coherence showed a negative Cliff’s delta, indicating reduced discourse-level organization in dementia speech.

\begin{figure}[h]
\centering
\includegraphics[width=0.9\linewidth]{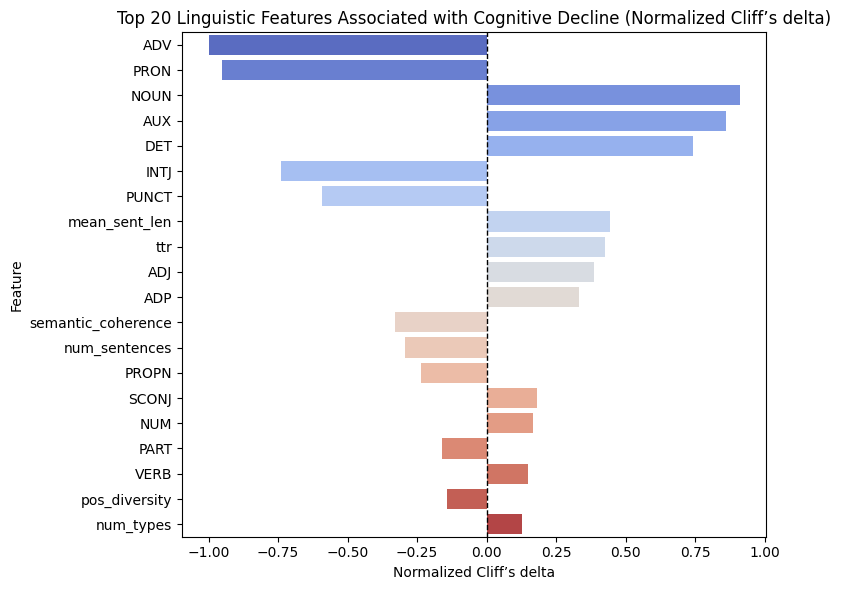}
\caption{Normalized Cliff’s delta values for the top linguistic features associated with cognitive decline. Negative values correspond to higher feature prevalence in dementia transcripts, while positive values indicate higher values in control transcripts. The magnitude of each bar reflects the strength of the non-parametric effect size, illustrating systematic differences in functional word usage, syntactic structure, and discourse-level features between the two groups.}
\label{fig:fig7}
\end{figure}

Figure~\ref{fig:fig7} visualizes the normalized Cliff’s delta values for the top features, illustrating both the direction and relative strength of group differences. Negative values correspond to features that are more prevalent in dementia transcripts, while positive values indicate higher values in control speech. The figure highlights a clear separation between functional, syntactic, and structural features associated with cognitive decline.

Overall, the statistical analysis supports the machine learning findings by identifying consistent linguistic markers of early cognitive decline that are robust to modeling assumptions. Features that emerged as influential in classification models, including pronouns, auxiliary verbs, determiners, adverbs, sentence length, and semantic coherence, also exhibited statistically significant group differences with small to moderate non-parametric effect sizes. This convergence between predictive modeling and statistical analysis strengthens the evidence that these linguistic characteristics reflect meaningful and systematic changes in language associated with early cognitive impairment.

\subsection{Subject-Level Cross-Validation Results}

\subsubsection{Classification Performance on Raw Cleaned Text}

To address concerns regarding potential information leakage arising from multiple transcripts per participant, classification performance was re-evaluated using subject-level cross-validation. A five-fold GroupKFold strategy was applied, ensuring that all transcripts from a given participant were assigned exclusively to either the training or the test set within each fold. This evaluation protocol provides a more conservative and clinically realistic estimate of model generalization to unseen speakers.

All results reported in this subsection are averaged across the five folds and presented together with standard deviations. Logistic regression and random forest models were trained using the same raw cleaned text features as in the transcript-level experiments. Hyperparameters were selected based on preliminary tuning and then fixed across all folds to ensure consistency. Logistic regression used L2 regularization with balanced class weights and a maximum of 1000 iterations, while the random forest model consisted of 200 trees with square-root feature sampling and balanced class weighting.

Table~\ref{tab:tab6} summarizes the subject-level classification performance. Logistic regression achieved a mean accuracy of 0.61 with a standard deviation of 0.08 and a macro-averaged F1-score of 0.60 $\pm$ 0.08. Precision and recall were well balanced across classes, with macro-averaged precision of 0.60 $\pm$ 0.07 and recall of 0.60 $\pm$ 0.07. These results indicate moderate but stable performance when evaluated under strict subject separation.

\begin{table*}[h]
\centering
\caption{Subject-level classification performance on raw cleaned text features using five-fold cross-validation (mean $\pm$ standard deviation)}
\label{tab:tab6}
\begin{tabular}{lcccc}
\hline
Model & Accuracy & Precision (Macro) & Recall (Macro) & F1-score (Macro) \\
\hline
Logistic Regression & $0.61 \pm 0.08$ & $0.60 \pm 0.07$ & $0.60 \pm 0.07$ & $0.60 \pm 0.08$ \\
Random Forest & $0.60 \pm 0.04$ & $0.59 \pm 0.04$ & $0.58 \pm 0.04$ & $0.58 \pm 0.04$ \\
\hline
\end{tabular}
\end{table*}

The random forest model achieved a mean accuracy of 0.60 $\pm$ 0.04 and a macro-averaged F1-score of 0.58 $\pm$ 0.04. Macro-averaged precision and recall were 0.59 $\pm$ 0.04 and 0.58 $\pm$ 0.04, respectively. While slightly lower than logistic regression, the random forest model demonstrated consistent performance across folds. As expected, overall performance under subject-level evaluation was lower than that observed under transcript-level splits, reflecting the increased difficulty of generalizing across speakers rather than individual speech samples.

\begin{figure}[h]
\centering
\includegraphics[width=0.9\linewidth]{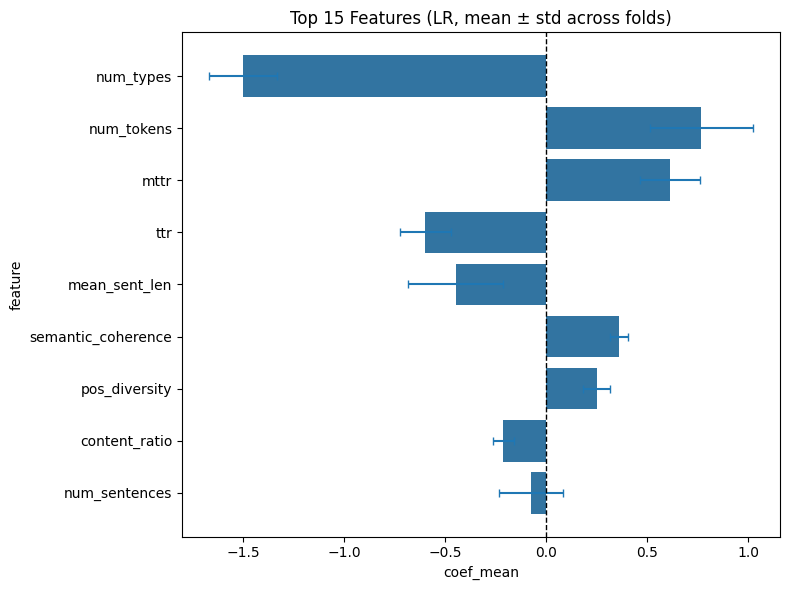}
\caption{Top logistic regression feature coefficients averaged across subject-level cross-validation folds (mean $\pm$ standard deviation)}
\label{fig:fig8}
\end{figure}

\begin{figure}[h]
\centering
\includegraphics[width=0.9\linewidth]{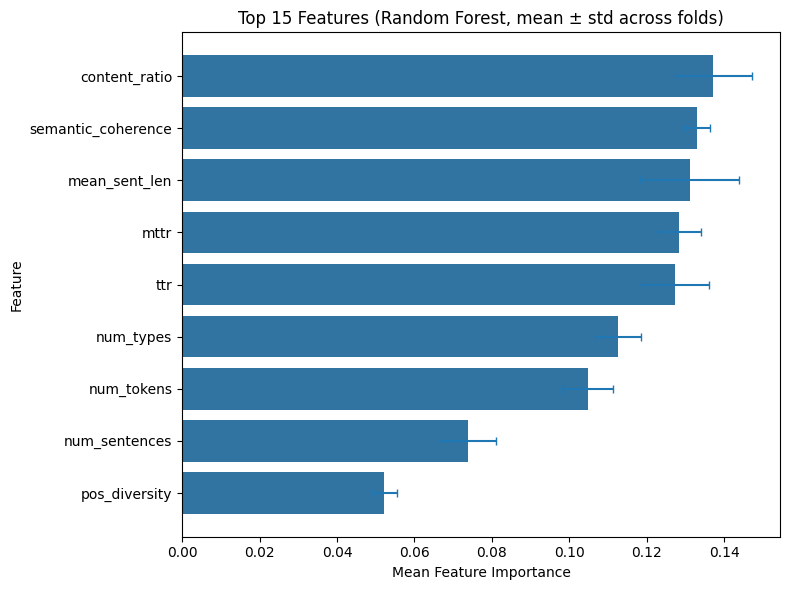}
\caption{Top random forest feature importances averaged across subject-level cross-validation folds (mean $\pm$ standard deviation).}
\label{fig:fig9}
\end{figure}

Feature importance analysis aggregated across cross-validation folds is shown in Figures \ref{fig:fig8} and \ref{fig:fig9}. For logistic regression, the most influential features included the number of word types, total number of tokens, lexical diversity measures such as type-token ratio and moving-average type-token ratio, and sentence-level structural features including mean sentence length and number of sentences. Semantic coherence and part-of-speech diversity also contributed consistently, indicating the relevance of discourse organization even in the absence of explicit syntactic tagging.

The random forest model exhibited a similar pattern of feature importance. Content word ratio, semantic coherence, mean sentence length, and lexical diversity measures emerged as the strongest predictors across folds, followed by structural measures such as the number of word types, total token count, and number of sentences. The stability of these feature rankings across folds suggests that the identified linguistic markers reflect robust group-level differences rather than artifacts of a particular data split.

Overall, the subject-level evaluation confirms that raw cleaned text features retain meaningful discriminatory power under a conservative and clinically realistic protocol. Although absolute performance decreases relative to transcript-level evaluation, the consistency of influential features across models and folds supports the robustness of lexical diversity, sentence structure, and discourse coherence as linguistic indicators of early cognitive decline.

\subsubsection{Classification Performance with POS-Enhanced Representation}

To further examine the contribution of syntactic information under a clinically realistic evaluation protocol, subject-level classification experiments were conducted using the POS-enhanced representation. As in the previous subsection, a five-fold GroupKFold strategy was employed to ensure that no participant appeared in both training and testing sets. Performance metrics are reported as the mean and standard deviation across folds.

Model hyperparameters were fixed across all folds following preliminary tuning. Logistic regression used L2 regularization with balanced class weights and a maximum of 1000 iterations, while the random forest model consisted of 200 trees with square-root feature sampling and balanced class weighting.

Using the POS-enhanced feature representation, logistic regression achieved a mean accuracy of 0.72 $\pm$ 0.07 and a macro-averaged F1-score of 0.71 $\pm$ 0.07. Macro-averaged precision and recall were both 0.71 $\pm$ 0.07, indicating balanced performance across classes (Table \ref{tab:tab7}). Compared to the raw-text subject-level experiments, the inclusion of part-of-speech information led to a substantial improvement in classification performance and reduced sensitivity to lexical variability across speakers.

The random forest model achieved a mean accuracy of 0.67 $\pm$ 0.08 and a macro-averaged F1-score of 0.66 $\pm$ 0.08. Precision and recall were 0.67 $\pm$ 0.08 and 0.66 $\pm$ 0.08, respectively (Table \ref{tab:tab7}). Although performance was lower than that of logistic regression, the random forest results remained stable across folds. Overall, both models benefited from the POS-enhanced representation, confirming that syntactic and grammatical cues provide complementary information beyond lexical content alone.

\begin{table*}[h]
\centering
\caption{Subject-level classification performance using POS-enhanced features (mean $\pm$ standard deviation across five folds)}
\label{tab:tab7}
\begin{tabular}{lcccc}
\hline
Model & Accuracy & Precision & Recall & F1-score \\
\hline
Logistic Regression & $0.72 \pm 0.07$ & $0.71 \pm 0.07$ & $0.71 \pm 0.07$ & $0.71 \pm 0.07$ \\
Random Forest & $0.67 \pm 0.08$ & $0.67 \pm 0.08$ & $0.66 \pm 0.08$ & $0.66 \pm 0.08$ \\
\hline
\end{tabular}
\end{table*}

\begin{figure}[h]
\centering
\includegraphics[width=0.9\linewidth]{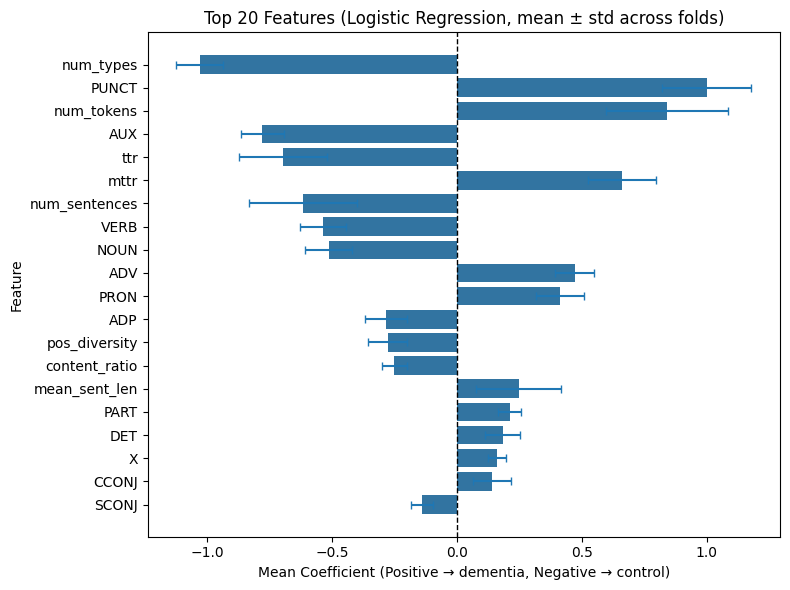}
\caption{Top logistic regression feature coefficients for the POS-enhanced representation, averaged across subject-level cross-validation folds (mean $\pm$ standard deviation). Positive coefficients indicate association with dementia transcripts.}
\label{fig:fig10}
\end{figure}

Feature importance aggregated across subject-level cross-validation folds provides insight into how lexical and syntactic features jointly contribute to classification decisions. For logistic regression, the most influential features included both lexical diversity measures and part-of-speech categories. As shown in Figure \ref{fig:fig10}, features such as the number of word types, type-token ratio, and moving-average type-token ratio remained prominent, alongside grammatical categories including punctuation, auxiliary verbs, pronouns, adverbs, and determiners. Structural features such as sentence length and the number of sentences also contributed, indicating that both grammatical usage and speech organization play a role in distinguishing between dementia and control transcripts.

\begin{figure}[h]
\centering
\includegraphics[width=0.9\linewidth]{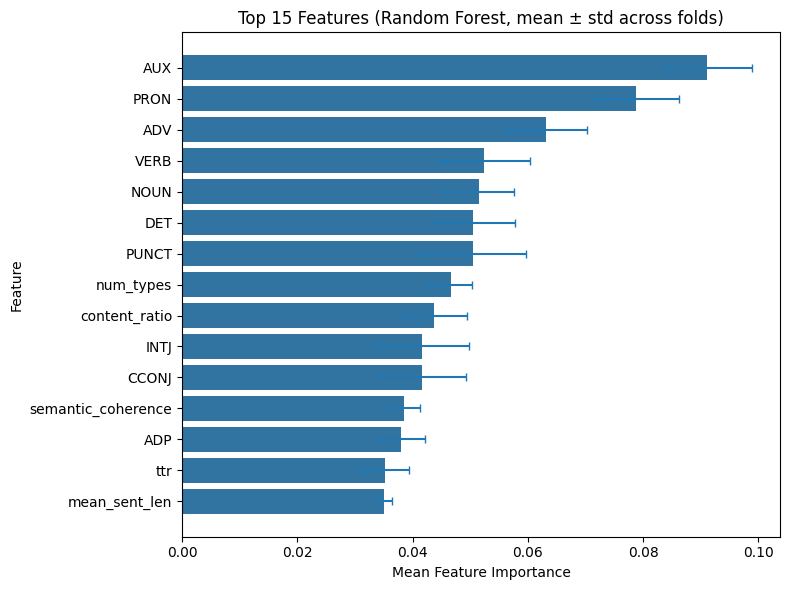}
\caption{Top random forest feature importances for the POS-enhanced representation, averaged across subject-level cross-validation folds (mean $\pm$ standard deviation).}
\label{fig:fig11}
\end{figure}

Feature importance results for the random forest model are shown in Figure \ref{fig:fig11}. In contrast to logistic regression, the random forest placed greater emphasis on part-of-speech features. Auxiliary verbs, pronouns, adverbs, verbs, nouns, and determiners emerged as the most influential predictors, followed by punctuation and lexical diversity measures. Semantic coherence and content word ratio also appeared among the top features, highlighting the continued importance of discourse-level organization even when syntactic information is explicitly encoded.

Across both models, there was substantial agreement in the types of features identified as important. Functional and grammatical categories, particularly pronouns, auxiliary verbs, adverbs, and determiners, consistently emerged as strong indicators of cognitive decline. The consistency of these findings across folds and across linear and non-linear models suggests that POS-enhanced representations capture stable and generalizable linguistic patterns associated with early cognitive impairment.

\subsubsection{Subject-Level Classification Performance with Full POS-Only Representation}
Finally, we evaluated subject-level classification performance using a POS-only representation in which lexical identity was removed and transcripts were represented exclusively through full part-of-speech distributions and structural linguistic features. Columns corresponding to lexical richness metrics that contained only missing values were excluded prior to training. This representation isolates syntactic and grammatical patterns, allowing us to assess their standalone discriminative capacity.

\begin{table*}[h]
\centering
\caption{Subject-level classification performance using POS-only features (mean $\pm$ standard deviation across five folds).}
\label{tab:tab8}
\begin{tabular}{lcccc}
\hline
\textbf{Model} & \textbf{Accuracy} & \textbf{Precision (macro)} & \textbf{Recall (macro)} & \textbf{F1-score (macro)} \\
\hline
Logistic Regression & $0.72 \pm 0.07$ & $0.71 \pm 0.07$ & $0.71 \pm 0.07$ & $0.71 \pm 0.07$ \\
Random Forest       & $0.67 \pm 0.06$ & $0.66 \pm 0.07$ & $0.66 \pm 0.07$ & $0.65 \pm 0.07$ \\
\hline
\end{tabular}
\end{table*}

\begin{figure}[h]
\centering
\includegraphics[width=0.9\linewidth]{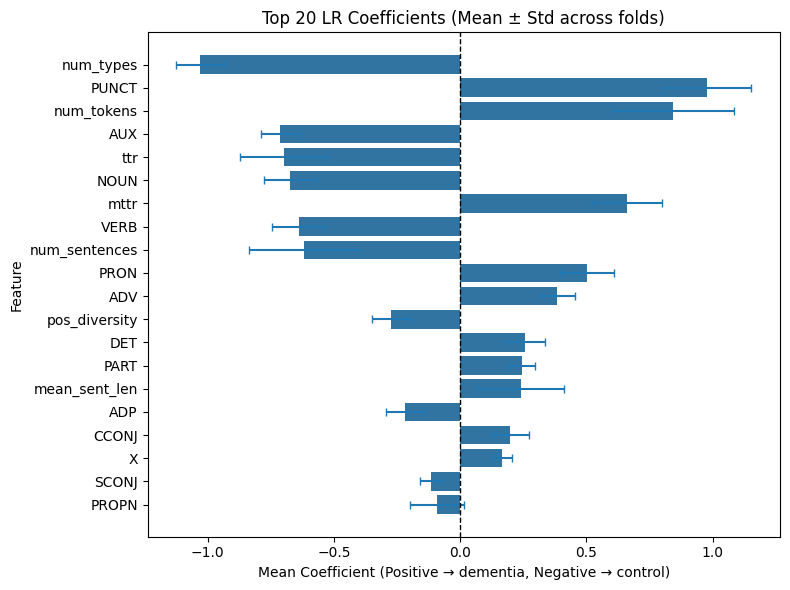}
\caption{Subject-level classification performance of the logistic regression model using the full POS-only representation under five-fold GroupKFold cross-validation. Results are reported as mean ± standard deviation across folds.}
\label{fig:fig12}
\end{figure}

Logistic regression achieved a mean accuracy of 0.72 with a macro-averaged F1-score of 0.71 under subject-level cross-validation (Table \ref{tab:tab8}). Precision and recall values were well balanced across folds, indicating stable classification behavior. These results closely match those obtained with the POS-enhanced representation, suggesting that grammatical structure alone captures a substantial portion of the linguistic signal associated with early cognitive decline, as shown in Figure~\ref{fig:fig12}.

\begin{figure}[h]
\centering
\includegraphics[width=0.9\linewidth]{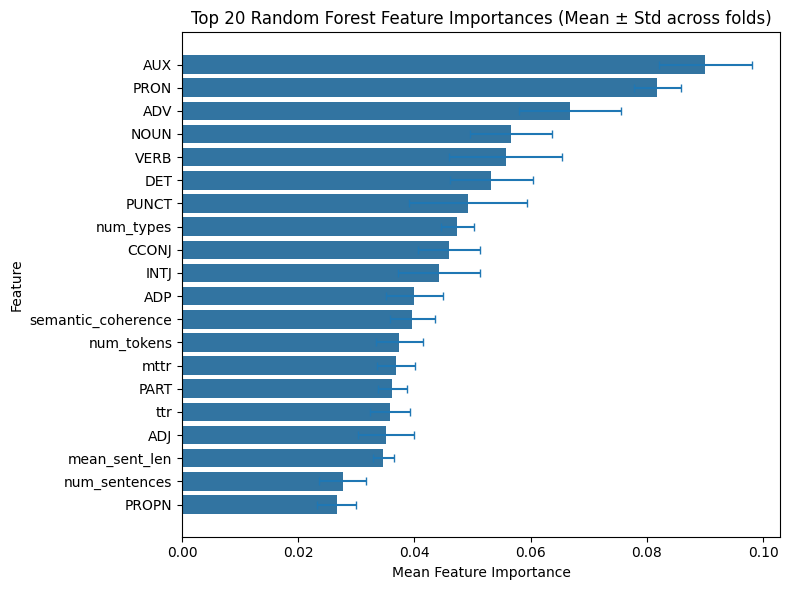}
\caption{Subject-level classification performance of the random forest model using the full POS-only representation under five-fold GroupKFold cross-validation. Results are reported as mean ± standard deviation across folds.}
\label{fig:fig13}
\end{figure}

The random forest model achieved a mean accuracy of 0.67 with a macro-averaged F1-score of 0.65 (Table \ref{tab:tab8}). While slightly lower than logistic regression, performance remained consistent across folds. Feature importance analysis, shown in Figure~\ref{fig:fig13}, revealed that auxiliary verbs, pronouns, adverbs, nouns, and determiners were among the most influential predictors. Disfluency-related markers such as punctuation and interjections also contributed meaningfully, highlighting disruptions in speech flow.

Across both models, the prominence of functional and grammatical categories supports the hypothesis that degradation in syntactic organization and reference management is a core linguistic characteristic of early cognitive decline. The comparable performance of POS-only and POS-enhanced representations under subject-level evaluation further suggests that syntactic patterns generalize more reliably across speakers than surface lexical content.

\subsubsection{Statistical Validation of Linguistic Markers under Subject-Level Evaluation}

To further validate the robustness of the linguistic indicators identified in the classification experiments, a subject-level statistical analysis was conducted using non-parametric tests and effect sizes. Table \ref{tab:tab9} reports the top linguistic features associated with early cognitive decline based on the Mann–Whitney U test, with effect sizes quantified using Cliff’s delta and p-values adjusted for multiple comparisons. Figure 14 provides a visual summary of these results by displaying normalized Cliff’s delta values for the top-ranked features, illustrating both the magnitude and direction of group differences.

The results reveal clear and consistent patterns across lexical, syntactic, and discourse-level features. Several part-of-speech categories exhibit statistically significant differences between dementia and control speech. Auxiliary verbs show the strongest positive effect size, indicating higher relative usage in control transcripts. In contrast, adverbs and pronouns display the largest negative effect sizes, reflecting increased reliance on modifiers and referential forms in dementia speech. These findings align with clinical observations that cognitive decline is often associated with reduced syntactic control and increased dependence on vague or less content-specific linguistic constructions.

Content-related grammatical categories such as nouns and determiners also show positive effect sizes, suggesting that control participants produce more content-dense and structurally anchored utterances. Conversely, higher rates of interjections and punctuation in dementia transcripts point to greater speech disruption and disfluency. Structural measures further differentiate the groups: control speech is characterized by longer sentences, a higher number of unique word types, and greater overall lexical variety, while dementia speech tends to be shorter, more repetitive, and less lexically diverse.

Discourse-level organization is also affected. Semantic coherence exhibits a negative effect size, indicating weaker topical continuity in dementia transcripts. Although its magnitude is smaller than that of some syntactic features, the effect remains statistically significant after correction, underscoring the role of discourse degradation as an early marker of cognitive decline.

Figure 14 highlights these trends by clearly separating features associated with higher values in control speech from those elevated in dementia speech. The consistency between the statistical analysis and the machine learning feature importance results strengthens the interpretation that these linguistic markers are not model-specific artifacts, but rather reflect stable and meaningful differences in language use associated with early cognitive impairment.

Overall, this subject-level statistical validation complements the predictive modeling results by providing independent evidence that changes in grammatical structure, functional word usage, lexical diversity, and discourse coherence are central characteristics of early cognitive decline. Together, these findings support the use of linguistically informed features as reliable indicators for automated dementia detection under clinically realistic evaluation settings.

\begin{table*}[h]
\centering
\caption{Top linguistic features associated with cognitive decline based on Mann--Whitney U test and Cliff’s delta}
\label{tab:tab9}
\begin{tabular}{lccccc}
\hline
Feature & Mean Control & Mean Dementia & Cliff’s $\delta$ & $p$-value & $p_{\text{adj}}$ \\
\hline
AUX & 0.0760 & 0.0592 & 0.357 & $5.70\times10^{-13}$ & $1.48\times10^{-11}$ \\
ADV & 0.0218 & 0.0345 & $-0.308$ & $5.04\times10^{-10}$ & $6.56\times10^{-9}$ \\
PRON & 0.1026 & 0.1213 & $-0.274$ & $3.27\times10^{-8}$ & $2.84\times10^{-7}$ \\
DET & 0.1278 & 0.1166 & 0.224 & $6.14\times10^{-6}$ & $3.35\times10^{-5}$ \\
NOUN & 0.1961 & 0.1810 & 0.223 & $7.09\times10^{-6}$ & $3.35\times10^{-5}$ \\
INTJ & 0.0131 & 0.0192 & $-0.221$ & $7.74\times10^{-6}$ & $3.35\times10^{-5}$ \\
PUNCT & 0.1121 & 0.1227 & $-0.219$ & $1.03\times10^{-5}$ & $3.83\times10^{-5}$ \\
ADP & 0.0892 & 0.0807 & 0.172 & $5.13\times10^{-4}$ & $1.67\times10^{-3}$ \\
num\_types & 64.20 & 57.94 & 0.164 & $9.18\times10^{-4}$ & $2.65\times10^{-3}$ \\
CCONJ & 0.0397 & 0.0469 & $-0.153$ & $1.96\times10^{-3}$ & $5.10\times10^{-3}$ \\
semantic\_coherence & 0.2344 & 0.2509 & $-0.143$ & $3.97\times10^{-3}$ & $9.37\times10^{-3}$ \\
VERB & 0.1437 & 0.1368 & 0.138 & $5.49\times10^{-3}$ & $1.16\times10^{-2}$ \\
mean\_sent\_len & 8.67 & 8.21 & 0.137 & $5.78\times10^{-3}$ & $1.16\times10^{-2}$ \\
num\_tokens & 114.18 & 105.65 & 0.111 & $2.50\times10^{-2}$ & $4.10\times10^{-2}$ \\
PROPN & 0.0148 & 0.0192 & $-0.108$ & $2.52\times10^{-2}$ & $4.10\times10^{-2}$ \\
PART & 0.0168 & 0.0193 & $-0.090$ & $6.80\times10^{-2}$ & $9.82\times10^{-2}$ \\
SCONJ & 0.0110 & 0.0091 & 0.089 & $6.45\times10^{-2}$ & $9.82\times10^{-2}$ \\
ADJ & 0.0297 & 0.0276 & 0.083 & $9.31\times10^{-2}$ & $1.27\times10^{-1}$ \\
num\_sentences & 13.39 & 13.08 & 0.059 & $2.32\times10^{-1}$ & $2.87\times10^{-1}$ \\
NUM & 0.0054 & 0.0052 & 0.056 & $2.00\times10^{-1}$ & $2.60\times10^{-1}$ \\
\hline
\end{tabular}
\end{table*}

\begin{figure}[h]
\centering
\includegraphics[width=0.9\linewidth]{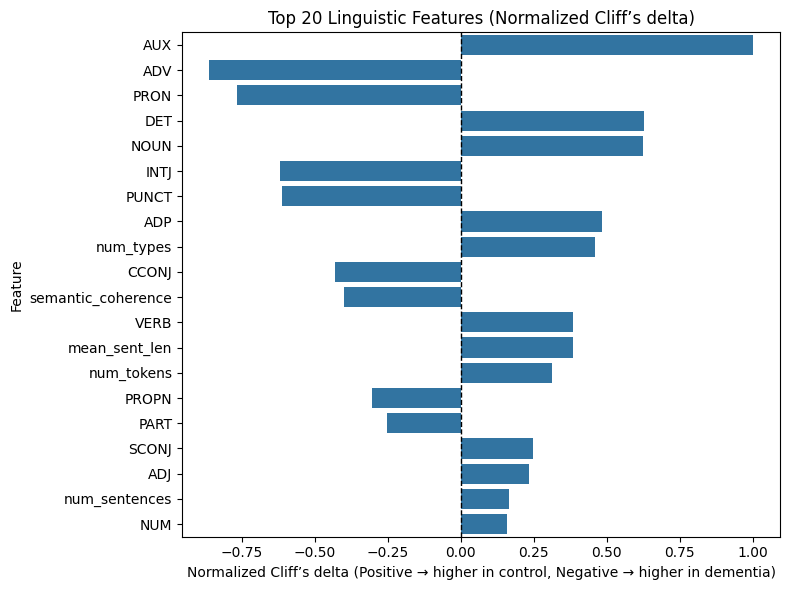}
\caption{Normalized Cliff’s delta values for the top linguistic features under subject-level evaluation. Positive values indicate features more prevalent in control speech, while negative values indicate higher prevalence in dementia speech. The magnitude reflects the strength of the non-parametric effect size.}
\label{fig:fig14}
\end{figure}

\subsection{Limitations and Future Directions}

Although the DementiaBank Pitt Corpus contains multiple recordings for some participants, the present study adopts a cross-sectional rather than longitudinal analysis. Longitudinal modeling of cognitive decline requires consistent and densely sampled recordings from the same individuals across time, which are not uniformly available in the corpus. As a result, this work focuses on identifying linguistic markers that distinguish early cognitive decline at the transcript level across individuals, rather than modeling within-subject progression. Future work could extend this framework to longitudinal settings where sufficient repeated measures per participant are available, enabling analysis of individual trajectories of linguistic change.

While the present study provides a detailed and interpretable analysis of linguistic markers associated with early cognitive decline, several limitations should be acknowledged. First, all experiments are conducted using the DementiaBank Pitt Corpus, with a strong emphasis on the Cookie Theft picture description task. Although this dataset is well established and widely used, its reliance on a limited set of elicitation tasks may constrain the generalizability of the findings to other conversational contexts, languages, cultural backgrounds, and education levels. As a result, the identified linguistic patterns should be interpreted as task and corpus-specific indicators rather than universal markers of cognitive decline.

Second, the proposed models rely exclusively on text-derived linguistic features and do not incorporate acoustic or prosodic information. Prior research suggests that speech-based markers such as pitch variation, speech rate, and pause structure can provide complementary signals for detecting cognitive impairment. Integrating linguistic and acoustic modalities represents a promising direction for future work, particularly if interpretability can be preserved.

Finally, language-based models for cognitive decline are inherently vulnerable to demographic and proxy biases, as well as uncertainty arising from heterogeneous clinical data. Recent methodological work has proposed uncertainty-aware and information-theoretic regularization strategies to mitigate such effects in neural models \cite{r25}. Although the present study focuses on classical, interpretable classifiers and does not explicitly model uncertainty or bias, incorporating uncertainty-aware bias mitigation techniques could improve robustness and fairness in future extensions of this framework.

All experiments in this study are conducted on the DementiaBank Pitt Corpus, with a strong emphasis on the Cookie Theft picture description task. While this task is widely used in clinical language assessment, it represents a constrained and culturally specific elicitation context. The findings may therefore not generalize directly to spontaneous conversational speech, other languages, or populations with different educational and cultural backgrounds. Evaluating the proposed features across diverse datasets and speech contexts remains an important direction for future research.

Future research may also explore longitudinal modeling of language change within individuals, cross-linguistic validation on non-English datasets, and clinician-centered evaluation of interpretability to better assess the practical utility of linguistic explanations in real-world clinical settings. 

Although this study emphasizes interpretability through global feature importance and statistical association analysis, interpretability is evaluated at a methodological level rather than through user-centered validation. The explanations presented here are intended to support linguistic insight and hypothesis generation, rather than to serve as decision support tools for clinicians. Future work should incorporate local explanation methods, stability analysis of explanations across resamples, and direct evaluation with clinicians to assess usability, trust, and clinical relevance.

The classification performance reported in this study is intentionally modest when compared to results reported by recent deep learning approaches. Many studies achieving accuracies above 90\% rely on acoustic–prosodic features, large pretrained language models, or transcript-level splits that allow speaker overlap between training and testing sets. In contrast, the present work adopts a subject-level evaluation protocol and focuses exclusively on linguistically interpretable features. This design choice prioritizes robustness and clinical transparency over maximal predictive accuracy. While modern transformer-based models may achieve higher performance, their representations are often opaque and difficult to translate into actionable linguistic insights for clinicians. The results of this study therefore reflect a deliberate tradeoff between accuracy and interpretability, rather than a limitation of the modeling approach.

\section{Conclusion}

This work examined linguistic indicators of early cognitive decline through a combination of machine learning and statistical analysis applied to speech transcripts from the DementiaBank Pitt Corpus. By evaluating multiple linguistic representations, including raw text, POS-enhanced features, and purely syntactic abstractions, the study demonstrates that meaningful signals of cognitive impairment persist even when lexical content is removed.
Importantly, results were analysed under two experimental protocols. Transcript-level experiments provide a point of comparison with prior studies, while subject-level cross-validation offers a more conservative and clinically realistic evaluation by preventing speaker overlap between training and testing data. Although subject-level performance is lower than transcript-level results, classification remains stable across models and feature sets, reinforcing the robustness of the identified linguistic markers.

While deep learning approaches reported in prior work often achieve accuracies above 85\% under transcript-level evaluation, such performance is frequently obtained without subject-level separation and with limited interpretability, making direct comparison to the clinically realistic protocol adopted here methodologically non-equivalent.

Feature importance analysis across logistic regression and random forest models consistently highlighted lexical diversity, functional word usage, sentence structure, and discourse coherence as key contributors to classification decisions. These findings were independently validated through non-parametric statistical testing using Mann–Whitney U tests and Cliff’s delta effect sizes. The statistical analysis revealed significant group differences in grammatical categories such as pronouns, auxiliary verbs, adverbs, and determiners, as well as in structural and discourse-level measures, confirming that the machine learning results reflect genuine linguistic differences rather than model-specific artifacts.
Together, these findings suggest that degradation in syntactic structure, functional word usage, and discourse organisation constitutes a core linguistic signature of early cognitive decline. The convergence of predictive modeling and statistical validation under subject-level evaluation strengthens the interpretability and reliability of the proposed approach. 

Future work may extend this framework to longitudinal analysis, cross-linguistic datasets, and multimodal speech features to further enhance early detection and clinical applicability. By demonstrating that abstract grammatical and syntactic features remain informative under subject-level evaluation and align with independent statistical evidence, this study supports the use of interpretable linguistic markers as a principled foundation for clinically grounded language-based screening.

\section*{Acknowledgment} This research used data from the DementiaBank Pitt Corpus. We acknowledge the team that created and provided access to this resource, which was supported by the National Institute on Aging (grants NIA AG03705 and AG05133).

\section*{Data Availability Statement} The data that support the findings of this study are available from the DementiaBank Pitt Corpus. Access to the dataset is restricted and requires registration and approval through DementiaBank (https://dementia.talkbank.org).

\section*{Conflict of Interest} The authors declare no conflict of interest.

\end{document}